\documentclass[10pt,journal,compsoc]{IEEEtran}
\usepackage{amsfonts,amsmath,amssymb,amsthm,bm}
\usepackage{graphicx,psfrag,epsf}
\usepackage{enumerate}
\usepackage{adjustbox}
\usepackage{dsfont}
\usepackage{comment}
\usepackage{xcolor}
\usepackage{multirow}
\usepackage{lipsum}

\usepackage{hyperref}
\usepackage{algorithm}
\usepackage{algpseudocode}
\algnewcommand{\Inputs}[1]{%
  \State \textbf{Inputs:}
  \Statex \hspace*{\algorithmicindent}\parbox[t]{.8\linewidth}{\raggedright #1}
}
\algnewcommand{\Initialize}[1]{%
  \State \textbf{Initialize:}
  \Statex \hspace*{\algorithmicindent}\parbox[t]{.8\linewidth}{\raggedright #1}
}

\newcommand{\R}{\mathbb{R}}

\ifCLASSOPTIONcompsoc
  \usepackage[nocompress]{cite}
\else
  \usepackage{cite}
\fi

\ifCLASSINFOpdf

\else
 
\fi

\ifCLASSOPTIONcompsoc
 \usepackage[caption=false,font=footnotesize,labelfont=sf,textfont=sf]{subfig}
\else
 \usepackage[caption=false,font=footnotesize]{subfig}
\fi

\begin{document}

\title{Deep Learning with Label Noise: \\A Hierarchical Approach}

\author{Li~Chen,~Ningyuan~(Teresa)~Huang,~Cong~Mu,~Hayden~S.~Helm,~Kate~Lytvynets, ~Weiwei~Yang,\\Carey E.~Priebe
% <-this % stops a space
\IEEEcompsocitemizethanks{
\IEEEcompsocthanksitem Li Chen is Research Scientist with Meta AI. E-mail: lichen66@fb.com 
 %\protect\\
\IEEEcompsocthanksitem Ningyuan (Teresa) Huang and Cong Mu are PhD students in the Department of Applied Mathematics and Statistics, Johns Hopkins University. E-mail: nhuang19@jhu.edu, cmu2@jhu.edu 
\IEEEcompsocthanksitem
Hayden S.\ Helm,
Kate Lytvynets, and
Weiwei Yang
are with Microsoft Research.
E-mail:
haydenshelm@gmail.com, kalytv@microsoft.com, weiwya@microsoft.com
\IEEEcompsocthanksitem Carey E.\ Priebe
is Professor in the Department of Applied Mathematics and Statistics (AMS), the Center for Imaging Science (CIS), and the Mathematical Institute for Data Science (MINDS), Johns Hopkins University. E-mail: cep@jhu.edu %\protect\\
\protect
}

}

\markboth{IEEE Transactions on Pattern Analysis and Machine Intelligence}%
{Shell \MakeLowercase{\textit{et al.}}: Bare Demo of IEEEtran.cls for Computer Society Journals}

\IEEEtitleabstractindextext{%
\begin{abstract} %first sentence tweak
Deep neural networks are susceptible to label noise. Existing methods to improve robustness, such as meta-learning and regularization, usually require significant change to the network architecture or careful tuning of the optimization procedure. In this work, we propose a simple hierarchical approach that incorporates a label hierarchy when training the deep learning models. Our approach requires no change of the network architecture or the optimization procedure. We investigate our hierarchical network through a wide range of simulated and real datasets and various label noise types. 
%, and can be applied to both supervised learning and transfer learning scenarios.
%, in which either a label hierarchy or a feature hierarchy is enforced. 
Our hierarchical approach improves upon regular deep neural networks in learning with label noise. Combining our hierarchical approach with pre-trained models achieves state-of-the-art performance in real-world noisy datasets.
% particularly for class-dependent label noise setting. 
\end{abstract}

%https://www.sciencedirect.com/science/article/pii/S0950705121000344#sec3

}

\maketitle

\IEEEdisplaynontitleabstractindextext

% \fi
\IEEEpeerreviewmaketitle

% \IEEEraisesectionheading{\section{Introduction}\label{sec:introduction}}

%\IEEEPARstart{C}{onsider} 

\section{Introduction}
\label{sec:introduction}

The robustness of deep learning has been studied from different aspects. One of the topics focuses on investigating whether deep neural networks can learn from noisy labels as it can be difficult to collect data with clean annotations in many real applications~\cite{algan2020label,song2022learning}. Although deep learning models enjoy certain generalizability for different tasks, they can be very sensitive to label noise as they tend to memorize noise during training due to their expressivity~\cite{zhang2021understanding,arpit2017closer}. In addition, there exist different types of label noise~\cite{frenay2013classification} and each of them may have its unique effects on the model performance.

Deep learning models designed to mitigate label noise can be broadly categorized into two groups: model-based and model-free \cite{algan2021image}. Model-based methods depend on explicit assumptions about the distribution and the behavior of the noise. Popular techniques in model-based settings include noisy channel~\cite{patrini2017making,hendrycks2018using,goldberger2016training}, data pruning~\cite{huang2019o2u,sharma2020noiserank,yan2016robust} and sample selection~\cite{jiang2018mentornet,han2018co,chen2019understanding}. Model-free methods, on the other hand, aim to improve robustness without explicitly modeling the label noise structure. Common model-free methods include robust losses such as non-convex loss~\cite{manwani2013noise,ghosh2015making,charoenphakdee2019symmetric}, generalized cross-entropy loss~\cite{zhang2018generalized}, meta-learning~\cite{garcia2016noise,li2017learning,algan2021meta}, and regularization~\cite{hendrycks2019using,jindal2016learning,azadi2015auxiliary} among others~\cite{cao2012noise,yu2018learning,duan2016learning}. We focus on the model-free setting in this paper due to its broad applicability.

\begin{figure}[h]
\centering
\includegraphics[width=8cm]{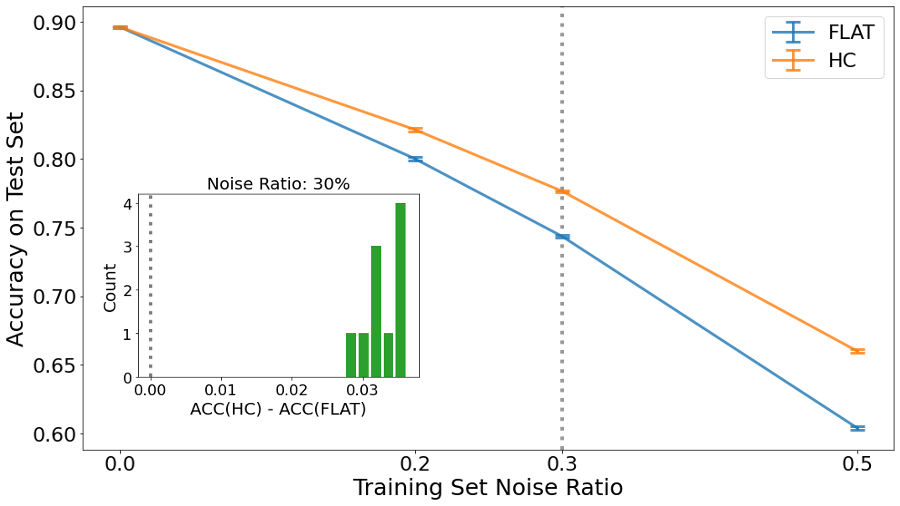}
\caption{Performance advantages obtained by our hierarchical model (HC) compared to the standard model (FLAT) on ICON94 dataset with uniform noise and noise ratio $ \in \left\{0\%, 20\%, 30\%, 50\% \right\} $.
The accuracy gain (of HC over FLAT) is both statistically and operationally significant.}
\label{fig:illustration}
\end{figure}

We propose a simple and efficient hierarchical approach that requires no change of the network architecture and the optimization mechanism. This is in contrast to the existing model-free methods above that typically require either significant change in network architecture (e.g., co-teaching in \cite{han2018co, yu2019does}) or in the optimization procedure (e.g., meta-learning in \cite{algan2021meta}, semi-supervised learning in \cite{ding2018semi, li2020dividemix}). Experiments on the benchmark datasets with synthetic noise suggest that our proposed hierarchical approach can statistically (and operationally) improve the performance of the deep learning model (see Figure \ref{fig:illustration}). This result is similarly observed in a real-world dataset with inherent label noise.

The rest of this paper is organized as follows. Section~\ref{sec:background} provides background on label noise taxonomy and related work on the main directions of this topic.
Section~\ref{sec:HCmodel} introduces our proposed hierarchical model. Section~\ref{sec:experiments} provides experimental results on different datasets with various types of label noise. Section~\ref{sec:ablation} conducts ablation study on our proposed method. Section~\ref{sec:discussion} discusses our findings and future work.

% Appendices provide extra details of our experiments.

% \begin{itemize}
%     \item Deep learning classifiers against label noise can be broadly categorized into two groups: noise model based and noise model free methods \cite{algan2021image}. 
%     \item In this work, we focus on noise model free methods, which aims to improve robustness without explicitly modeling the label noise structure.
%     \item Common noise model free methods include robust losses, meta-learning, and regularization.
%     \item We propose a hierarchical approach that requires no change of the network architecture and can be easily adapted in transfer learning scenario.
% \end{itemize}
\section{Background} \label{sec:background}

\subsection{Label Noise Taxonomy} \label{subsec:taxonomy}
In this section, we provide different major categorizations of label noise. The simplest type of label noise is known as uniform label noise, where the ground truth labels are changed to the wrong labels uniformly with a probability $p$. A slightly more complicated type of label noise is class-dependent. In class-dependent settings there is a noise transition matrix $ T_{K \times K} $, where $ K $ is the number of classes, that governs the probability of a ground-truth label getting switched to a different label. In particular, if an observation has a ground truth label $ i $ then the probability that the observed class label is $ j $ is $ T_{ij} $. Hence, the diagonal of the matrix $ T $ is proportion of truly-class $ i $ remaining class $ i $. The transition matrix $ T $ need not be symmetric. A yet more complicated setting is feature and class-dependent. That is, the probability of transitioning from class $ i $ to class $ j $ is a function of both the feature vector and the ground truth label $ i $. This setting best mimics real-world label noise scenarios as the difficulty in labeling for human annotators is non-uniform for most ground-truth class conditionals. Some authors consider part-dependent label noise where the noise only partially depends on an instance \cite{xia2020part}. 

Also related to the label noise taxonomy is the perspective of \cite{chen2021} and \cite{goel2021robustness} where they describe the label noise problem in the language of uncertainty. In their characterization, there are two types of uncertainty: aleatoric and epistemic. Aleatoric uncertainty persists in the data even as the number of samples goes to infinity whereas epistemic uncertainty can be avoided with a sufficient amount of data. The uniform label noise setting we consider herein falls under aleatoric uncertainty, and the class-dependent noise setting under both aleatoric and epistemic uncertainty.

\subsection{Related work} \label{subsec:related_work}
Major directions of mitigating label noise include developing robust architecture, selecting samples with clean labels during training, modifying the loss function to estimate the inverse of the transition matrix, and leveraging deep learning uncertainty techniques. Authors in \cite{ghosh2017robust, zhang2018generalized, wang2019symmetric, lyu2019curriculum} proposed a robust loss function to achieve a smaller risk for unseen clean data. Sample selection techniques that filter clean labels for training and remove noisy labels have been proposed in \cite{jiang2018mentornet, han2018co, yu2019does, malach2017decoupling}. Sample selection and label correction for spatial computing are studied in \cite{chen2020robust}.
Devising loss to estimate the noise transition matrix and correct the labels are studied in \cite{patrini2017making, hendrycks2018using, arazo2019unsupervised}. Various uncertainty techniques to mitigate label noise are proposed, such as using MCDropout \cite{chen2021, goel2021robustness}, Bootstrap \cite{bootstrap}, and Bayesian CNN using Bayes by Backprop \cite{10.5555/3045118.3045290}. Semi-supervised learning is another emerging field in label noise research, where the noisy labeled data are treated as unlabeled \cite{nguyen2019self, ding2018semi, li2020dividemix}. 

\section{Deep Hierarchical Model}

\label{sec:HCmodel}

\subsection{Problem Setup}

Consider the classical classification set up \cite[pg.2]{dgl}: Let $(X,Y), (X_1,Y_1), \cdots (X_n,Y_n) \overset{iid}{\sim} F_{XY}$,
where a feature vector realization $X$ is an element of $\mathbb{R}^d$ and class label $Y$ realization is an element of $[K] = \{1,\cdots,K\}$.
Denote the training data by $\mathcal{T}_n = \{(X_1,Y_1), \cdots (X_n,Y_n)\}$.
Our goal is to learn a classifier $g: \R^d \times (\R^d \times [K])^n \to [K]$ using $\mathcal{T}_n$
  to predict the true but unobserved class label $Y$ based on the observed test feature vector $X$.
Performance is measured by conditional probability of error,
\begin{equation}
    L(g) = P_F[g(X;\mathcal{T}_n) \neq Y|\mathcal{T}_n]. \label{eqn:error}
\end{equation}

Now consider the setting where we do not observe the $Y_i$ but rather noisy labels $Z_i$.
% symmetric noise simulation scheme:
% noise transition matrix:
% for the ith label, the KxK matrix with (1-P_i) on the diagonal and P_i/(K-1) on the off-diagonal; that is, M = (P_i/(K-1))*J + ((1-Pi))-(P_i/(K+1))*I where J is the KxK all 1s matrix and I is the KxK identity matrix.
For $P_i \in [0,1]$,
 let noisy class label $Z_i$ be given by
 $P[Z_i = Y_i] = 1-P_i$
 and $Z_i$ distributed uniformly on $[K] \setminus \{Y_i\}$ with probability $P_i$;
 $P_i = 0$ means no noise in the $Z_i$ and $P_i = (K-1)/K$ means no information in the noisy labels $Z_i$.
Thus, we have
$(X_i,Y_i,Z_i,P_i) \overset{iid}{\sim} F_{X,Y,Z,P}$.
Again: $X$ is the feature vector and $Y$ is the true class label;
now $Z$ is the noisy class label and $P$ characterizes the label noise. The classifier $g$ is trained on the noisy dataset $\tilde{\mathcal{T}}_n = \{(X_1,Z_1), \cdots (X_n,Z_n)\}$, and evaluated on the clean sample
\begin{equation}
    L(\tilde{g}) = P_F[g(X;\tilde{\mathcal{T}}_n) \neq Y|\tilde{\mathcal{T}}_n]. \label{eqn:error}
\end{equation}

The goal is to find $\tilde{g} = g(X; \tilde{\mathcal{T}}_n)$ such that $L(\tilde{g})$ is minimized. In what follows, we consider $g_{\theta}$ as a deep neural network parameterized by $\theta$.% Our approach can be applied to either $g_{\theta}$ is trained from scratch or $g_{\theta}$ is pre-trained from some other datasets.

%\subsection{Overview}

%We view the classifier $f$ obtained from clean data is composed of an encoder $g$ that learns a (feature) representation, and a linear classifier $l$ that decides the labels. Namely, $f = l \circ g$. Let $f'$ be the noisy classifier trained from the noisy dataset. Then $f'$ can differ from $f$ in the encoder or/and the classifier. Transfer learning in $f'$ --- freezing the encoder and finetuning the classifier --- preserves robustness in the encoder, which can be seen as a form of meta-learning. Hierarchical models improve robustness against label noise, which take forms of robust losses and regularization. We connect transfer learning to a feature hierarchy, and hierarchical classification to a label hierarchy. See Figure~\ref{fig:overview} for an overview of our proposed hierarchical training scheme.
 
\subsection{Hierarchical Model}
For some multi-class classification problems, the labels can be structured into a hierarchy. A hierarchy may be based on an expert's definition of what similarity means, e.g., labels corresponding to the concept of ``mammal" such as ``tiger", ``human", and ``lion" may all share a subsection of the label tree. Or, a hierarchy may be based on the shared features of different classes, e.g., ``dolphin" and ``shark" may share a subsection of the label tree. In hierarchical classification settings, the coarser label (i.e., ``mammal") can be viewed as a more robust classification signal than the original fine class labels because mislabels more naturally occur within a subsection of the hierarchy -- it is easier to confuse a tiger for a lion than a tiger for a shark. In this work, we utilize the label hierarchy to improve classification performance in the presence of label noise.

\begin{figure}[h]
\centering
\includegraphics[width=8.5cm]{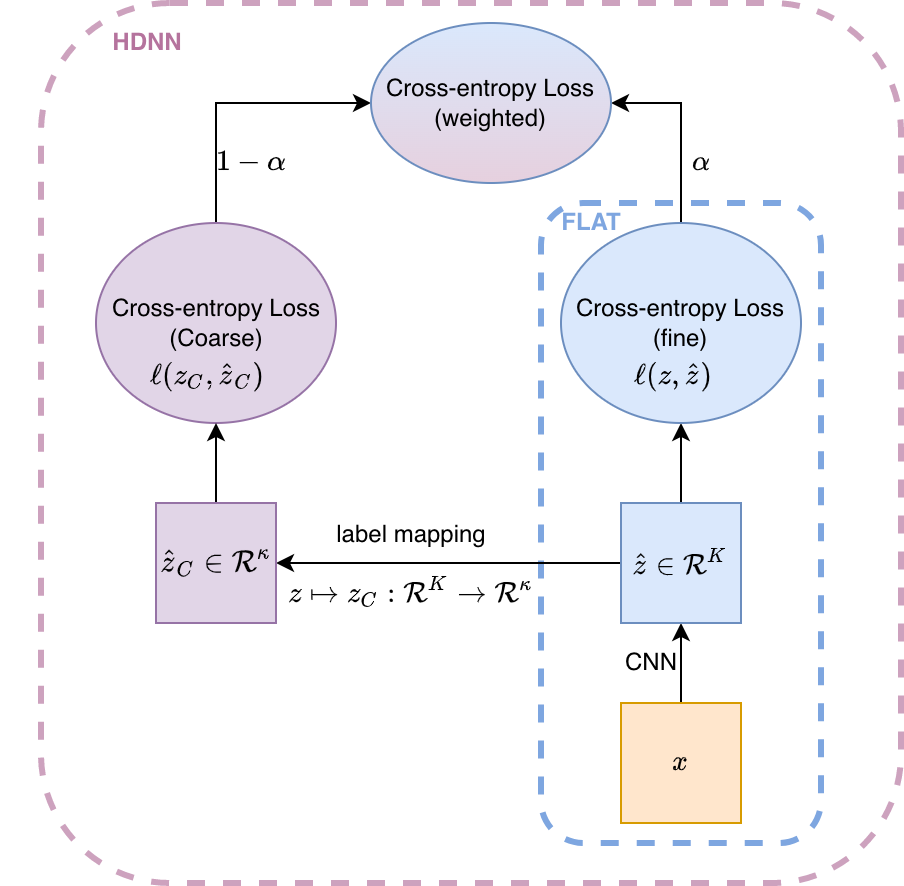}
\caption{Our proposed hierarchical (HC) model compared to the standard (FLAT) model: HC model only requires a label mapping function and a weighted loss objective in addition to the original FLAT model. Our HC model runs as fast as the FLAT model and reduces to the FLAT model when $\alpha = 1$.}
\label{fig:overview}
\end{figure}

More precisely: Let $\hat{z}, \hat{z}_C \in \mathbb{R}^K$ be the predicted probability of the original class label and the coarse class label, respectively. Let $\mathbf{\hat{z}},\mathbf{\hat{z}_C} \in \mathbb{R}^{n \times K}$ be the predictions arranged in rectangular form. Let $\mathbf{x} \in \R^{n \times d}$ be the design matrix, and $\mathbf{z} \in \R^{n \times K}$ be the noisy labels (in one-hot encoding form). Algorithm 1 and Figure \ref{fig:overview} describe our hierarchical approach. The key idea is to augment the standard neural network $g_{\theta}$ with a label hierarchy mapping function $f$ and a weighted loss function $L$, without any changes of the network architecture. Figure~\ref{fig:illustration} illustrates the performance advantages obtained by our hierarchical approach (details and elaboration are provided in Section 4.3 below).
 
\begin{algorithm}
\caption{Hierarchical Model}\label{alg:hdnn}
\begin{algorithmic}
\Require Label mapping function $f$, weight scalar $\alpha$, model $g_{\theta}$, data $(\mathbf{x}, \mathbf{z})$, epochs $T$, learning rate $\eta$. %, node features $X \in \R^{n \times k}$

\Initialize{$\theta$ from pre-trained model or randomly.} % = X
\For{t = 0 to T}
\State{$\mathbf{\hat{z}} = g_{\theta}(\mathbf{x})$ \, [Forward pass]} 
\State{$\mathbf{\hat{z}_C} = f(\mathbf{\hat{z}})$} \, [Coarse label mapping]
\State{$L = \left(1-\alpha \right) \ell(\mathbf{\hat{z}_C}, \mathbf{{z}_C}) + \alpha \ell(\mathbf{\hat{z}}, \mathbf{{z}})$ \, [Weighted Loss]} 
\State{$\theta = \theta - \eta \nabla_{\theta}L$ } \, [Backward pass]
\EndFor \\
\Return $g_{\theta}$
\end{algorithmic}
\end{algorithm}

As shown in Figure~\ref{fig:overview}, we construct a label hierarchy using a label mapping function, which maps the predictions of the original classes (fine) to the higher-level classes (coarse). The label mapping function can be given a priori or can be learned from the data ~\cite{helm2021inducing}. We will discuss this in more details in Section~\ref{sec:experiments}. 

% \begin{equation}
%     L = \left(1-\alpha \right) \ell_{\text{coarse}} + \alpha \ell_{\text{fine}},
% \end{equation}
Once we have mapped coarse label predictions, we compute the weighted loss function per Algorithm \ref{alg:hdnn}, where the weight $ \alpha $ can be adjusted to deal with different noise ratios. Note that when $\alpha = 1$, our HC model reduces back to the original FLAT model. In practice, one can fix $ \alpha $ based on any prior knowledge of data quality or adaptively adjust $ \alpha $ during the training process. See Section~\ref{sec:ablation} for the investigation of $ \alpha $ on the performance of our proposed method.

Notably, our HC model has the same order of computational complexity as the original FLAT model: per Algorithm \ref{alg:hdnn}, the HC model only requires an additional computation of the [Coarse Label Mapping] step, whereas all the other steps are the same as in the FLAT model. In contrast, many existing hierarchical frameworks increase the complexity of the network architecture (e.g., \cite{zhu2017b} adds higher-level classifiers) or the optimization procedure (e.g., \cite{yan2015hd} requires training on subnetworks for fine classes and coarse classes, and then fine-tuning end-to-end). Thus, our HC model is a simple, effective strategy against label noise that can be efficiently implemented in any existing neural network model without incurring significant additional computational costs.

\section{Experiments}

\label{sec:experiments}

We conduct several sets of experiments on different datasets with various types of label noise to compare the performance of our HC models against the FLAT models.

\subsection{Datasets}

We consider 3 datasets with synthetic noise and 1 real world noisy dataset, which are summarized as follows.

\begin{itemize}
    \item \textbf{MNIST}~\cite{lecun1998gradient}. The collection of handwritten digits (10-class) with a training set of 60000 examples and a test set of 10000 examples.
    \item \textbf{CIFAR100}~\cite{krizhevsky2009learning}. Labeled subset (100-class) of the 80 million tiny images dataset with a training set of 50000 examples and a test set of 10000 examples.
    \item \textbf{ICON94}. Subset (94-class) of the \href{https://www.kaggle.com/datasets/testdotai/common-mobile-web-app-icons}{Common Mobile/Web App Icons} with a training set of 113314 examples , a validation set of 14164 examples and a test set of 14164 examples. 
    \item \textbf{ANIMAL-10N}~\cite{song2019selfie}. A real world noisy (noise ratio around 8\%) dataset (10-class) with a training set of 50000 examples and a test set of 5000 examples.
\end{itemize}

\subsection{Setup}

\noindent
\textbf{Label Noise.}
We consider the following two types of synthetic noise in our experiments.

\begin{itemize}
    \item \textbf{Uniform Noise.} The probability of label changing from the true class to any other class is equally distributed, i.e, the noise transition matrix $ T $ is defined as
    \begin{equation}
        T_{ij} =
        \begin{cases}
            1-p & i = j \\
            \frac{p}{K-1} & \text{otherwise}
        \end{cases},
    \end{equation}
    where $ K $ is the number of class. In our experiments, we consider the noise ratio $ p \in \left\{0.2, 0.3, 0.5 \right\} $.
    \item \textbf{Class-dependent Noise.} The probability of label changing to other class depends on the true class of the data instance. In particular, we follow the procedure in~\cite{algan2020label} where one first train a deep neural network using training set, and then construct the noise transition matrix with the confusion matrix of the trained network on the test set. Note that the noise generated by this procedure is also feature-dependent since the noise transition matrix depends on the network trained on the features. In our experiments, we consider the noise ratio $ p \in \left\{0.25, 0.35, 0.45, 0.55 \right\} $. For MNIST and CIFAR100 datasets, we used the class-dependent noisy labels generated by ~\cite{algan2020label} that are made publicly available at \href{https://github.com/gorkemalgan/corrupting_labels_with_distillation}{this repository}.
\end{itemize}

\noindent
\textbf{Label Hierarchical Structure.} In practice, one can either extract the natural hierarchical structure from the data or learn the hierarchical structure as in~\cite{helm2021inducing}. The hierarchical structures we apply in our experiments are summarized as follows.

\begin{itemize}
    \item \textbf{MNIST}: the higher level (coarse) label $ Y_c $ (5-class) is constructed from the lower level (fine) label $ Y_f $ (10-class) as
    \begin{equation}
        Y_c = 
        \begin{cases}
            0 & Y_f \in \left\{0, 6 \right\} \\
            1 & Y_f \in \left\{1, 7 \right\} \\
            2 & Y_f \in \left\{2, 8 \right\} \\
            3 & Y_f \in \left\{3, 5 \right\} \\
            4 & Y_f \in \left\{4, 9 \right\}
        \end{cases}
    \end{equation}
    This is a natural hierarchical structure from the data as each of the pair of $ \left(0, 6 \right), \left(1, 7 \right), \left(2, 8 \right), \left(3, 5 \right), \left(4, 9 \right) $ is relatively similar in original images as handwritten digits.
    \item \textbf{CIFAR100}: the higher level (coarse) label $ Y_c $ (20-class) is constructed from the lower level (fine) label $ Y_f $ (100-class) following the nature of the data as the 100 classes in this dataset are already grouped into 20 superclasses.
    \item \textbf{ICON94}: the higher level (coarse) label $ Y_c $ (20-class) is constructed from the lower level (fine) label $ Y_f $ (94-class) using the methods described in~\cite{helm2021inducing}.
    \item \textbf{ANIMAL-10N}: the higher level (coarse) label $ Y_c $ (5-class) is constructed from the lower level (fine) label $ Y_f $ (10-class) as
    \begin{equation}
        Y_c = 
        \begin{cases}
            0 & Y_f \in \left\{\text{cat, lynx} \right\} \\
            1 & Y_f \in \left\{\text{jaguar, cheetah} \right\} \\
            2 & Y_f \in \left\{\text{wolf, coyote} \right\} \\
            3 & Y_f \in \left\{\text{chimpanzee, orangutan} \right\} \\
            4 & Y_f \in \left\{\text{hamster, guinea pig} \right\}
        \end{cases} \label{eqn:animal}
    \end{equation}
    This is a natural hierarchical structure from the data as each pair of (cat, lynx), (jaguar, cheetah), (wolf, coyote), (chimpanzee, orangutan), (hamster, guinea pig) looks similar and can lead to confusion for human annotator.
\end{itemize}

\noindent
\textbf{Architecture.}
To test the compatibility of our HC models with different network architectures, we adopt different backbones for each dataset. Specifically, \texttt{AlexNet}~\cite{krizhevsky2014one} is used for experiments on \textbf{MNIST}, \texttt{ResNet-18}~\cite{he2016deep} is used for experiments on \textbf{CIFAR100}, and \texttt{ResNeXt-50(32×4d)}~\cite{xie2017aggregated} is used for experiments on \textbf{ICON94} and \textbf{ANIMAL-10N}.

\hspace{1em}

\noindent
\textbf{Training Scheme.} 
The transformation for the input images follow the recommended steps for pre-trained models in \texttt{PyTorch}~\cite{paszke2019pytorch}. The batch size is set to be 64 and the networks (both FLAT and HC models) are trained for 100 epochs for all experiments. ADAM \cite{kingma2014adam} is used for optimization with the initial learning rate of $ 0.0001 $, which decays by a rate of 0.5 every 50 epochs. The weight parameter $ \alpha $ is set to be 0.5 for the first set of experiments. The effect of $ \alpha $ on performance is studied in Section \ref{sec:ablation}.

\subsection{Results}

To fully understand the performance of our HC models against FLAT models in the presence of label noise, we consider two metrics: per-epoch accuracy gain and final-stage accuracy gain. For per-epoch accuracy gain, we compare their test accuracies at each epoch and check the significance of the improvement of our HC models against FLAT models via McNemar’s test~\cite{dietterich1998approximate}; The per-epoch analysis shows how the HC models react to the label noise relative to the FLAT models during training and informs the optimal choice of early stopping; For final-stage accuracy gain, we compare the averaged test accuracies over the last 10 epochs of the HC and the FLAT models, which can be thought as the worst-case scenario when the label noise has fully contaminated the models. This can help us understand how well the HC models can mitigate the memorization of noisy labels relative to the FLAT models.

%This is useful since in practice one may adopt certain early stopping strategies. 
\hspace{1em}

\begin{figure}
	\centering
	\subfloat[Uniform noise with noise ratio 20\%, 30\% and 50\%.]{\includegraphics[width=0.48 \textwidth]{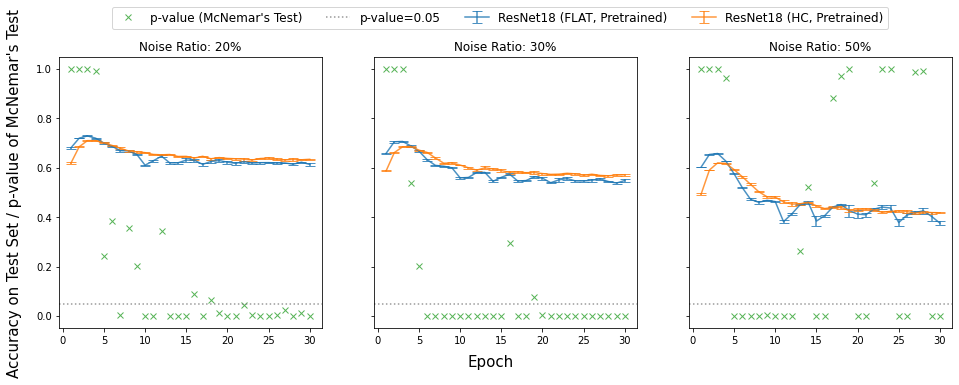}%
	}
	\hfil
	\subfloat[Class-dependent noise with noise ratio 25\%, 35\% and 55\%.]{\includegraphics[width=0.48 \textwidth]{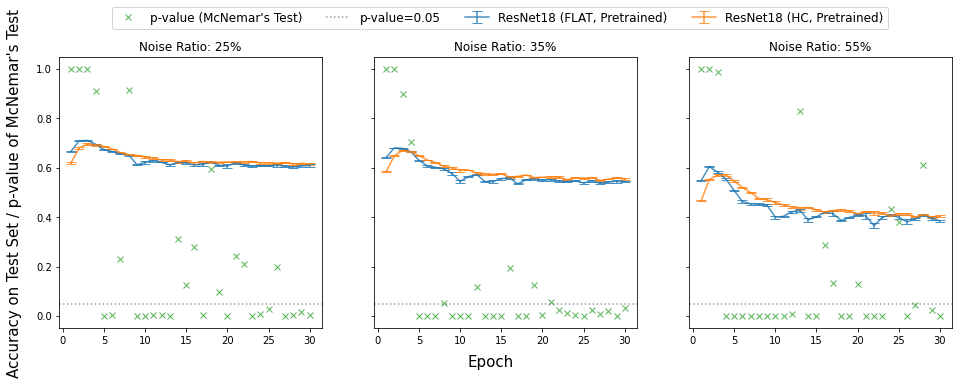}%
	}
	\caption{Per-epoch comparison of FLAT and HC models on CIFAR100 with synthetic noise. Test accuracy is plotted as \texttt{mean($\pm$stderr)} across 5 runs. \texttt{p-value} is obtained from the McNemar's test and plotted as \texttt{median} across 5 runs. HC is statistically significant better than FLAT (p-value smaller than $0.05$) for most of the epochs. }
	\label{fig:cifar100}
\end{figure}

\begin{figure}
	\centering
	\includegraphics[width=0.48\textwidth]{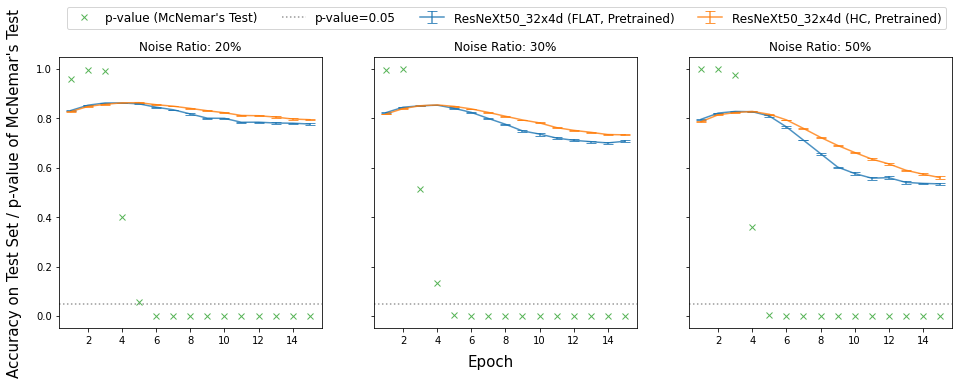}
	\caption{Per-epoch comparison of FLAT and HC models on ICON94 with uniform noise and noise ratio 20\%, 30\% and 50\%. Test accuracy is plotted as \texttt{mean($\pm$stderr)} across 5 runs. \texttt{p-value} is obtained from the McNemar's test and plotted as \texttt{median} across 5 runs. HC is statistically significant better than FLAT (p-value smaller than $0.05$) across all epochs larger than $5$ at different noise ratio levels. }
	\label{fig:icon94}
\end{figure}

\begin{figure}
\centering
\includegraphics[width=8cm]{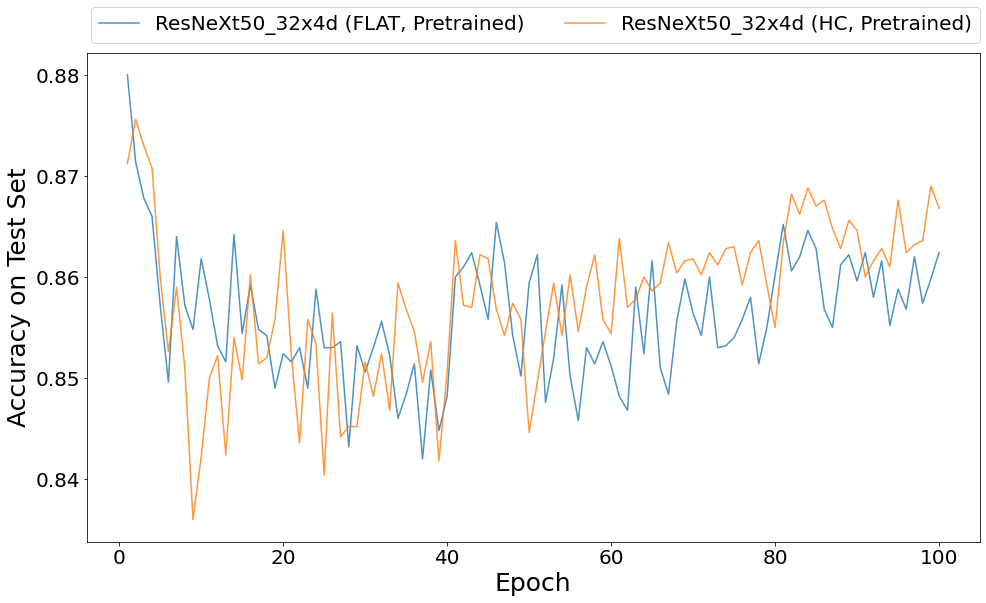}
\caption{Per-epoch comparison of FLAT and HC models on ANIMAL-10N. HC tends to outperform FLAT across epoch $50-100$.}
\label{fig:animal10n}
\end{figure}

\noindent
\textbf{Per-epoch accuracy gain.}
Figure~\ref{fig:cifar100} provides the per-epoch comparison of FLAT and HC models on \textbf{CIFAR100} with various types of synthetic noise and noise ratios. Figure~\ref{fig:icon94} provides the per-epoch comparison of FLAT and HC models on \textbf{ICON94} with uniform noise and different noise ratios. Note that our HC models perform significantly better than the FLAT models for most of epochs. Figure~\ref{fig:animal10n} provides the per-epoch comparison of FLAT and HC models on \textbf{ANIMAL-10N} of which the noise ratio is around 8\%. Again the proposed HC models have better performance in considerably many regions of the regime.

\hspace{1em}

\noindent
\textbf{Final-stage accuracy gain.} 
Table~\ref{tab:experiments_last10} provides comparison of FLAT and HC models in terms of their final-stage accuracies (i.e. averaged over epoch 90 to epoch 100). Table~\ref{tab:experiments_earlystopping} includes comparison of FLAT and HC models where performance is evaluated in terms of the test accuracy averaged over epoch 21 to epoch 30 for \textbf{MNIST} and \textbf{CIFAR100}, epoch 6 to epoch 15 for \textbf{ICON94}, which corresponds to the scenario when (optimal) early stopping is applied. Note that our HC models perform significantly better than the FLAT models for most of the experiments, including low-noise (even zero noise) and high-noise regimes; the advantage of HC models becomes more prominent when coupled with early stopping.

\begin{table}
\scriptsize
\caption{Comparison of FLAT and HC models on different datasets. Test accuracy is averaged over last 10 epochs and reported as \texttt{mean($\pm$stderr)} across 5 runs for synthetic noise.}
\label{tab:experiments_last10}
\centering
\begin{tabular}{ccccc}
\hline \hline
Dataset                   & Noise                     & Ratio & FLAT & HC  \\ \hline \hline
\multirow{6}{*}{MNIST}    & \multirow{3}{*}{Uniform}         & 20\%        & 96.42    $\pm$ 0.01 &  \textbf{96.48 $\pm$ 0.00}     \\
                         &                                  & 30\%        &92.34  $\pm$ 0.01 &  \textbf{92.59 $\pm$ 0.02}    \\
                          &                                  & 50\%        & \textbf{78.34  $\pm$0.02 } &78.06  $\pm$ 0.04      \\ \cline{2-5} 
                          & \multirow{3}{*}{Class-dependent} & 25\%        & 92.51 $\pm$  0.02  & \textbf{92.63 $\pm$ 0.02 }       \\ 
                          &                                  & 35\%        & 84.03 $\pm$  0.05  &   \textbf{84.26 $\pm$ 0.01 }      \\
                          &                                  & 45\%        & 71.71  $\pm$  0.06 & \textbf{72.41 $\pm$ 0.05 }    \\ \hline
\multirow{6}{*}{CIFAR100} & \multirow{3}{*}{Uniform}         & 20\%        & 63.25 $\pm$ 0.25 & \textbf{64.15 $\pm$ 0.13}     \\
                          &                                  & 30\%        & 56.23 $\pm$ 0.25 & \textbf{57.60 $\pm$ 0.24}     \\
                          &                                  & 50\%        & 39.43 $\pm$ 0.23 & \textbf{41.56 $\pm$ 0.11}     \\ \cline{2-5} 
                          & \multirow{3}{*}{Class-dependent} & 25\%        & 62.15 $\pm$ 0.06 & \textbf{62.55 $\pm$ 0.05}     \\
                          &                                  & 35\%        & 55.39 $\pm$ 0.13 & \textbf{55.99 $\pm$ 0.10}     \\
                          &                                  & 55\%        & 40.03 $\pm$ 0.12 & \textbf{40.85 $\pm$ 0.10}          \\ \hline
\multirow{3}{*}{ICON94}   & \multirow{3}{*}{Uniform}         & 20\%        &   79.31 $\pm$ 0.13   &  \textbf{79.77 $\pm$ 0.12}        \\
                          &                                  & 30\%        &   72.91 $\pm$ 0.02   &  \textbf{73.37 $\pm$ 0.19}        \\
                          &                                  & 50\%        &  55.58 $\pm$ 0.14    &  55.77 $\pm$ 0.20       \\ \hline
ANIMAL-10N                & Real                             & 8\%         &  85.94      & \textbf{86.38}           \\ \hline \hline
\end{tabular}
\end{table}

\begin{table}
\scriptsize
\caption{Comparison of FLAT and HC models using early stopping on different datasets. Test accuracy is averaged over epoch 21 to epoch 30 for MNIST and CIFAR100, epoch 6 to epoch 15 for ICON94 and reported as \texttt{mean($\pm$stderr)} across 5 runs.}
\label{tab:experiments_earlystopping}
\centering
\begin{tabular}{ccccc}
\hline \hline
Dataset                   & Noise                       &  Ratio & FLAT & HC  \\ \hline \hline
\multirow{7}{*}{MNIST}    & Clean         & 0\%        &   99.43  $\pm$ 0.00 &  \textbf{99.45 $\pm$ 0.00}        \\ \cline{2-5} 
                          & \multirow{3}{*}{Uniform}         & 20\%        &    96.32 $\pm$ 0.11 &  \textbf{96.82 $\pm$ 0.06}        \\
                         &                                  & 30\%        & 92.69 $\pm$ 0.16    &  \textbf{93.50 $\pm$ 0.09}         \\
                          &                                  & 50\%        &  77.84 $\pm$ 0.17  &  \textbf{79.58 $\pm$ 0.34}       \\ \cline{2-5} 
                          & \multirow{3}{*}{Class-dependent} & 25\%        & 91.38 $\pm$  0.25  & \textbf{92.30 $\pm$ 0.44 }         \\ 
                          &                                  & 35\%        & 83.02 $\pm$  0.47 &   \textbf{84.94 $\pm$ 0.78 }        \\
                          &                                  & 45\%        & 72.04  $\pm$  0.66 & 73.49 $\pm$ 0.88        \\  \hline
\multirow{7}{*}{CIFAR100} & Clean         & 0\%        &   75.52  $\pm$ 0.06  &  \textbf{75.76 $\pm$ 0.09}        \\ \cline{2-5} 
                          & \multirow{3}{*}{Uniform}         & 20\%        & 61.89 $\pm$ 0.10 & \textbf{63.41 $\pm$ 0.20}     \\
                          &                                  & 30\%        & 54.72 $\pm$ 0.18 & \textbf{57.19 $\pm$ 0.10}     \\
                          &                                  & 50\%        & 41.27 $\pm$ 0.15 & \textbf{42.23 $\pm$ 0.07}     \\ \cline{2-5} 
                          & \multirow{3}{*}{Class-dependent} & 25\%        & 60.85 $\pm$ 0.13 & \textbf{61.94 $\pm$ 0.09}     \\
                          &                                  & 35\%        & 54.39 $\pm$ 0.12 & \textbf{55.65 $\pm$ 0.10}     \\
                          &                                  & 55\%        & 39.38 $\pm$ 0.10 & \textbf{41.01 $\pm$ 0.16}          \\ \hline 
\multirow{4}{*}{ICON94}   & Clean         & 0\%        &     89.62 $\pm$ 0.06  &  89.61 $\pm$ 0.05       \\ \cline{2-5} 
                          & \multirow{3}{*}{Uniform}         & 20\%        &   80.02 $\pm$ 0.11   &  \textbf{82.14 $\pm$ 0.11}        \\
                          &                                  & 30\%        &   74.26 $\pm$ 0.18   &  \textbf{77.68 $\pm$ 0.07}        \\
                          &                                  & 50\%        &  60.41 $\pm$ 0.11    &  \textbf{66.01 $\pm$ 0.13}      \\ \hline \hline
\end{tabular}
\end{table}

\hspace{1em}

To understand how our HC models can mitigate label noise, we use the experiments on \textbf{ANIMAL-10N} as an illustration. Recall that this dataset contains 5 pairs of confusing animals (see eqn \ref{eqn:animal}). In other words, the original (fine) 10-class labels are relatively noisy but the higher-level (coarse) labels are relatively clean. Also recall the weighted loss we adopt as
\begin{equation}
    L = \left(1-\alpha \right) \ell_{\text{coarse}} + \alpha \ell_{\text{fine}},
\end{equation}
where $ \alpha = 1 $ for FLAT models and $ \alpha < 1 $ for HC models. Thus our HC models can outperform the FLAT models by leveraging the information from the relatively clean coarse labels.

\begin{figure}
\centering
\includegraphics[width=8cm]{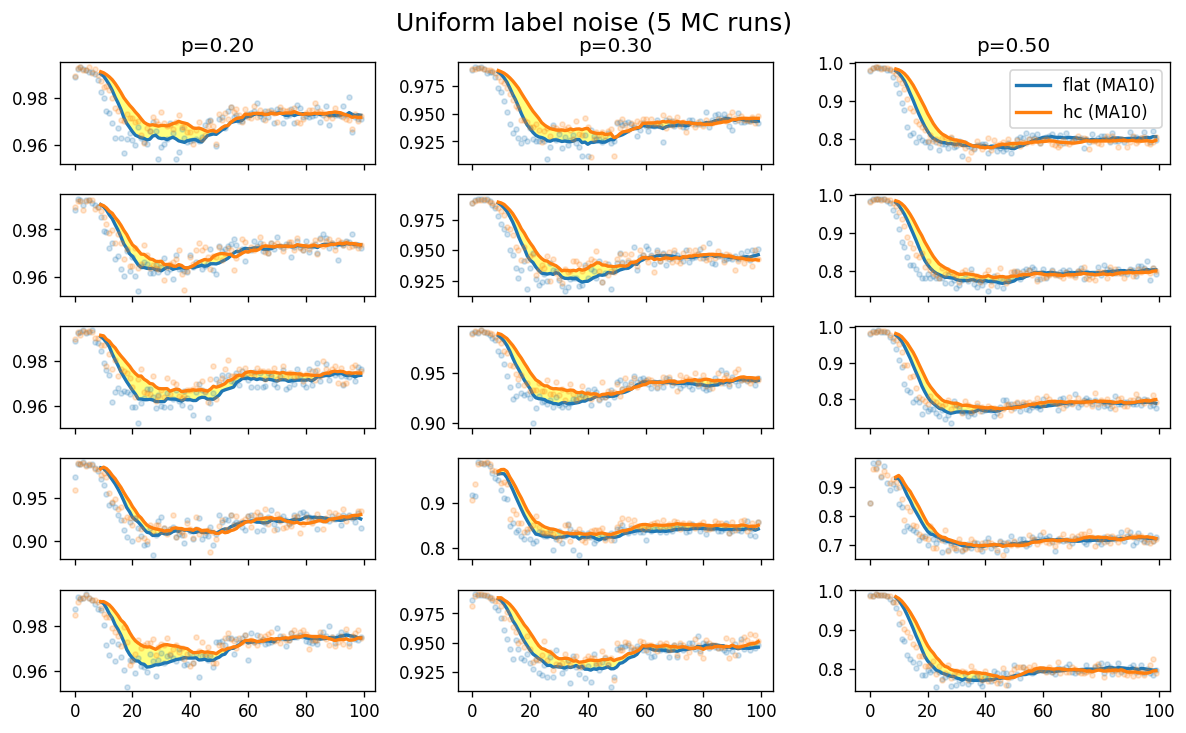}
\caption{Per-epoch comparison of FLAT and HC models on MNIST with uniform label noise. HC tends to outperform FLAT across epoch $10-50$, while converges to FLAT at epochs larger than $50$.}
\label{fig:mnist-per-epoch}
\end{figure}

\section{Ablation Study}

\label{sec:ablation}

% Recall the weighted loss we adopt is defined as
% \begin{equation}
%     L = \left(1-\alpha \right) \ell_{\text{coarse}} + \alpha \ell_{\text{fine}}.
% \end{equation}

To understand the effect of weight parameter $ \alpha $ on our proposed method, we conduct ablation study using \textbf{CIFAR100} with different types of noise and noise ratios.

\hspace{1em}

\noindent
\textbf{Setup.}
We consider uniform and class-dependent noise with different level of noise ratios, i.e., clean (0\%), low (20\%), medium (50\%), high (80\%), and weight parameter $ \alpha \in \left\{0.25, 0.5, 0.75, 1 \right\} $. Note that $ \alpha = 1 $ corresponds to the FLAT models. The model will leverage more information from the higher level (coarse) label as $ \alpha $ decreases. We use the same label hierarchical structure described in Section~\ref{sec:experiments}.

\hspace{1em}

\noindent
\textbf{Training Scheme.} 
\texttt{ResNet-18}~\cite{he2016deep} is used as the backbone network. The transformation for the input images follow the recommended steps for pre-trained models in \texttt{PyTorch}~\cite{paszke2019pytorch}. The batch size is set to be 64 and the networks are trained for 30 epochs for all experiments. ADAM \cite{kingma2014adam} is used for optimization with the learning rate of $ 0.0001 $.

\hspace{1em}

\begin{figure*}
	\centering
	\subfloat[Uniform noise with noise ratio 0\%, 20\%, 50\% and 80\%.]{\includegraphics[width=7in]{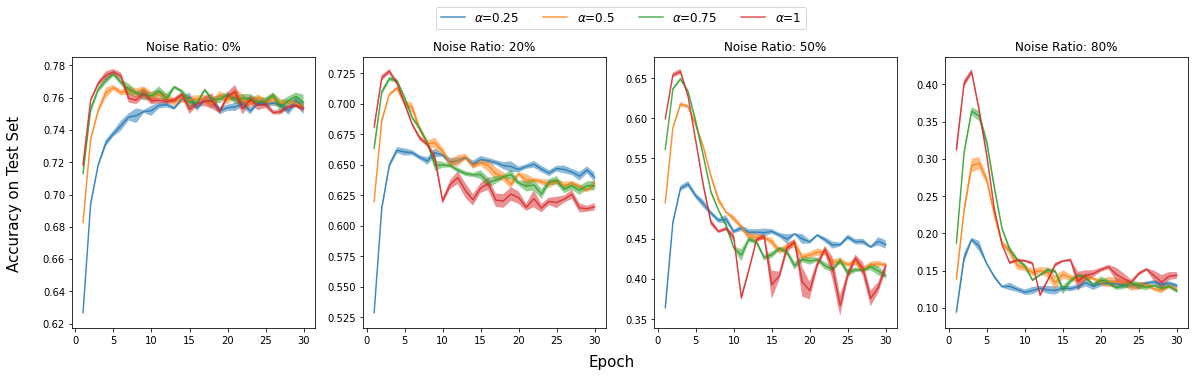}%
	}
	\hfil
	\subfloat[Class-dependent noise with noise ratio 0\%, 25\%, 55\% and 85\%.]{\includegraphics[width=7in]{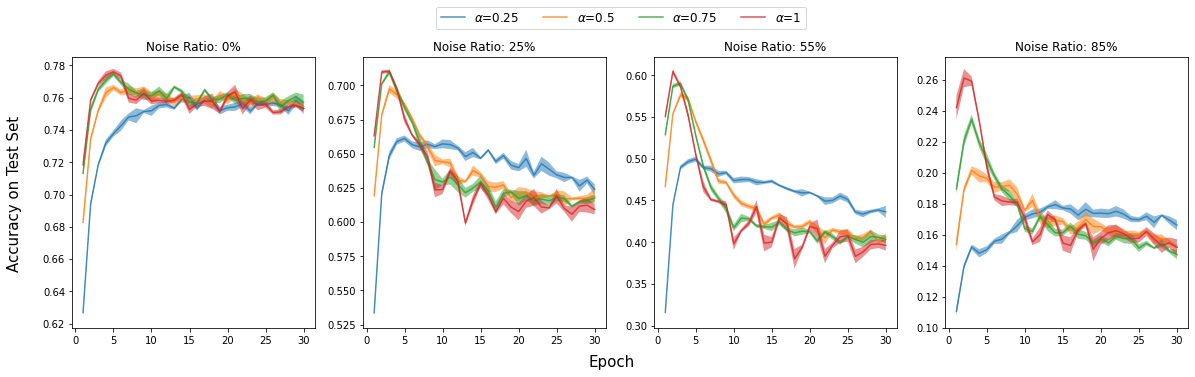}%
	}
	\caption{Per-epoch analysis of HC models with different weight parameter $ \alpha $ on CIFAR100 with synthetic noise. Test accuracy is plotted as \texttt{mean($\pm$stderr)} across 5 runs. The overall performance of HC models appear to be consistent for different weight parameter $ \alpha $.}
	\label{fig:ablation}
\end{figure*}

\noindent
\textbf{Results.} 
Figure~\ref{fig:ablation} provides per-epoch analysis of HC models with different weight parameter $ \alpha $ on \textbf{CIFAR100} with synthetic noise. Table~\ref{tab:experiments_ablation} includes comparison of HC models with different weight parameter $ \alpha $ on \textbf{CIFAR100} in which performance is evaluated in terms of the test accuracy averaged over epoch 21 to epoch 30. The overall performance of HC models appears to be consistent for different choices of $ \alpha $. In practice, we recommend using $\alpha = 0.5$. We leave the investigation of optimizing $\alpha$ for future work. %Future work will look into how to best select $ \alpha $ in label noise settings but we think that in practice $ \alpha = 0.5 $  is a safe choice.
% and one can further decrease $ \alpha $ to improve the performance for datasets that are not highly contaminated by the label noise.

\begin{table}
% \scriptsize
\tiny
\caption{Comparison of HC models with different weight parameter $ \alpha $ on CIFAR100. Test accuracy is averaged over epoch 21 to epoch 30 and reported as \texttt{mean($\pm$stderr)} across 5 runs.}
\label{tab:experiments_ablation}
\centering
\begin{tabular}{cccccc}
\hline \hline
\multirow{2}{*}{Noise}     &  \multirow{2}{*}{Ratio} & \multicolumn{4}{c}{$\alpha$}  \\ \cline{3-6} 
& & 0.25 & 0.5 & 0.75 & 1 \\ \hline \hline
Clean     & 0\%    &   75.50 $\pm$ 0.05     &     75.80 $\pm$ 0.09 &  75.84 $\pm$ 0.04   & 75.51  $\pm$ 0.06  \\ \cline{1-6} 
\multirow{3}{*}{Uniform}        
                                 & 20\%        & \textbf{64.51 $\pm$ 0.11} & 63.45 $\pm$ 0.14 & 63.21 $\pm$ 0.10 & 61.82 $\pm$ 0.15    \\
                                                            & 50\%        & \textbf{44.63 $\pm$ 0.13} & 42.14 $\pm$ 0.09 & 41.34 $\pm$ 0.20 & 40.60 $\pm$ 0.24     \\
                                                           & 80\%        & 13.17 $\pm$ 0.23 & 12.96 $\pm$ 0.28 & 12.89 $\pm$ 0.24 & \textbf{14.36 $\pm$ 0.30}     \\ \cline{1-6} 
                          \multirow{3}{*}{Class-dependent} & 25\%        & \textbf{63.43 $\pm$ 0.07} & 61.86 $\pm$ 0.19 & 61.61 $\pm$ 0.04 & 61.22 $\pm$ 0.14    \\
                                                           & 55\%        & \textbf{44.42 $\pm$ 0.14} & 40.88 $\pm$ 0.09 & 40.45 $\pm$ 0.13 & 39.72 $\pm$ 0.27    \\
                                                           & 85\%        & \textbf{17.10 $\pm$ 0.08} & 15.90 $\pm$ 0.12 & 15.37 $\pm$ 0.07 & 15.79 $\pm$ 0.12         \\ \hline \hline
\end{tabular}
\end{table}

\section{Discussion}

\label{sec:discussion}

To improve the robustness of deep learning models against label noise, we propose a hierarchical training scheme that utilizes a hierarchical label structure and a weighted loss objective. Compared with other methods which usually require significant change to the network architecture or careful tuning of the optimization procedure, our hierarchical approach is simple and accessible in the sense that we require no change of the network architecture or the optimization mechanism. Experiments on datasets with synthetic noise and real-world noisy datasets suggest that our proposed HC models can significantly improve the performance of the original FLAT models when training with label noise. To that end, one can adopt the existing state-of-the-art label noise algorithms with our hierarchical training scheme to further boost the model performance against label noise.

Our general hierarchical approach opens the door to many interesting future directions. Firstly, we can tailor the hierarchical approach for each specific type of label noise, such as using a feature hierarchy for feature-dependent label noise. Moreover, our current hierarchical classifier has no prior knowledge of the presence of noisy labels; its performance may be further improved by incorporating such priors, such as an option to discard certain labeled samples. Finally, we demonstrate the hierarchical classifier is not only more robust against label noise, but also generalizes well or even outperforms the flat model without label noise; investigating its generalization property both in-distribution and out-of-distribution is a natural next step.

\section*{Acknowledgements}

The authors thank Youngser Park for his valuable comments on the paper. Cong Mu and Teresa Huang are partially supported by the Johns Hopkins Mathematical Institute for Data Science (MINDS) Data Science Fellowship.

% \appendices

% \section{Detailed Experimental Results}

% \label{appendix:A}

% Figure~\ref{fig:cifar100_all_un} provides per-epoch comparison of FLAT and HC models on \textbf{CIFAR100} with uniform noise and different noise ratio for 5 runs. Figure~\ref{fig:cifar100_all_cdn} provides per-epoch comparison of FLAT and HC models on \textbf{CIFAR100} with class-dependent noise and different noise ratio for 5 runs. Figure~\ref{fig:icon94_all_un} provides per-epoch comparison of FLAT and HC models on \textbf{ICON94} with uniform noise and different noise ratio for 5 runs.

% \begin{figure*}
% 	\centering
% 	\includegraphics[width=6in]{Cong/CIFAR100_UN_p020305_all5runs_per-epoch.png}%
% 	\caption{Per-epoch comparison of FLAT and HC models on CIFAR100 with uniform noise for 5 runs.}
% 	\label{fig:cifar100_all_un}
% \end{figure*}

% \begin{figure*}
% 	\centering
% 	\includegraphics[width=6in]{Cong/CIFAR100_CDN_p025035055_all5runs_per-epoch.png}%
% 	\caption{Per-epoch comparison of FLAT and HC models on CIFAR100 with class-dependent noise for 5 runs.}
% 	\label{fig:cifar100_all_cdn}
% \end{figure*}

% \begin{figure*}
% 	\centering
% 	\includegraphics[width=6in]{Cong/ICON94_UN_p020305_all5runs_per-epoch.png}%
% 	\caption{Per-epoch comparison of FLAT and HC models on ICON94 with uniform noise for 5 runs.}
% 	\label{fig:icon94_all_un}
% \end{figure*}

% \clearpage
\bibliographystyle{IEEEtran}

% \bibliography{sample}
\bibliography{hierarchy, Li/Li_bib}

\end{document}